\documentclass[conference]{IEEEtran}
\IEEEoverridecommandlockouts
\usepackage{cite}
\usepackage{amsmath,amssymb,amsfonts}
\usepackage{algorithmic}
\usepackage{graphicx}
\usepackage{textcomp}
\usepackage{xcolor}
\usepackage{comment}
\usepackage{multirow}
\usepackage{makecell}
\DeclareUnicodeCharacter{0301}{\'{e}}
\def\BibTeX{{\rm B\kern-.05em{\sc i\kern-.025em b}\kern-.08em
    T\kern-.1667em\lower.7ex\hbox{E}\kern-.125emX}}
\begin{document}

\title{Real-Time Cattle Interaction Recognition via
Triple-stream Network\\
}
\author{\IEEEauthorblockN{1\textsuperscript{st} Yang Yang}
\IEEEauthorblockA{\textit{Graduate School of System Informatics} \\
\textit{Kobe University}\\
Kobe, Japan \\
yo@cs25.scitec.kobe-u.ac.jp}
\and
\IEEEauthorblockN{2\textsuperscript{nd}Mizuka Komatsu}
\IEEEauthorblockA{\textit{Graduate School of System Informatics} \\
\textit{Kobe University}\\
Kobe, Japan \\
m-komatsu@bear.kobe-u.ac.jp}
\and
\IEEEauthorblockN{3\textsuperscript{rd}Takeno Ohkawa}
\IEEEauthorblockA{\textit{Graduate School of System Informatics} \\
\textit{Kobe University}\\
Kobe, Japan \\
ohkawa@kobe-u.ac.jp}
\and
\IEEEauthorblockN{4\textsuperscript{th} Kenji Oyama}
\IEEEauthorblockA{\textit{Graduate School of Agricultural Science} \\
\textit{Kobe University}\\
Kobe, Japan \\
oyama@kobe-u.ac.jp}
}

\maketitle

\begin{abstract}
In stock breeding of beef cattle, computer vision-based approaches have been widely employed to monitor cattle conditions (e.g. the physical, physiology, and health). To this end, the accurate and effective recognition of cattle action is a prerequisite. Generally, most existing models are confined to individual behavior that uses video-based methods to extract spatial-temporal features for recognizing the individual actions of each cattle. However, there is sociality among cattle and their interaction usually reflects important conditions, e.g. estrus, and also video-based method neglects the real-time capability of the model. Based on this, we tackle the challenging task of real-time recognizing interactions between cattle in a single frame in this paper. The pipeline of our method includes two main modules: Cattle Localization Network and Interaction Recognition Network. At every moment, the cattle localization network outputs high-quality interaction proposals from every detected cattle and feeds them into the interaction recognition network with a triple-stream architecture. Such a triple-stream network allows us to fuse different features relevant to recognizing interactions. Specifically, the three kinds of features are a visual feature that extracts the appearance representation of interaction proposals, a geometric feature that reflects the spatial relationship between cattle, and a semantic feature that captures our prior knowledge of the relationship between the individual action and interaction of cattle. In addition, to solve the problem of insufficient quantity of labeled data, we pre-train the model based on self-supervised learning. Qualitative and quantitative evaluation evidences the performance of our framework as an effective method to recognize cattle interaction in real time.
\end{abstract}

\section{Introduction}
In recent years, with the development of AI technologies and the improvement of computer hardware performance, AI technologies have been widely used in many types of fields such as business, medicine, education, and agriculture, which have also been applied to animal husbandry to optimize feeding management. Traditionally, such management is implemented by manual observation. However, this method requires significant time and effort but with low accuracy. Sensor technologies are able to automatically monitor action patterns have been used for cattle\cite{b1} but some of these sensors are harmful and could cause stress to cattle. In addition, use of sensors also causes additional costs. Recently, livestock action recognition combines visual object detection with image classification, and has been the subject of increased interest in the fields of computer vision and smart agriculture\cite{b2, b3}. 

Computer vision technology, which enables non-attached observation of cattle, has recently attracted attention for action recognition. This research field usually focuses on monitoring cattle conditions such as estrus through recognizing the action like `riding'\cite{b4,b8}. The study of recognizing action has achieved great success by deep learning approaches \cite{b5,b6,b7} and among these approaches, video based recognition is relatively well-established and well-studied area of research, whereas still image based recognition is less studied. The main reason is that action recognition in still image suffer from the loss of spatio-temporal features. However, this conclusion is made totally based on human-centric experiments and since the number of cattle in the frame is generally big in the frame, extracting spatio-temporal features for them at the same time will lead to a significant increase in computation and affect the real-time performance of the system. By observing the behavioral characteristics of cattle, we find that many classes of cattle actions can be explicitly described in a still image without spatio-temporal features \textit{e.g., `grazing' can be recognized simply by characteristics of head}. This evidence supports us in using single frame for recognition to achieve real-time monitoring of cattle. Moreover, based on our observations, several salient regions can obviously represent the cattle's action. For instance, the characteristics of head can help us to recognize whether they are grazing, and the body part can obviously show whether they are lying or standing. Concentrating on such regions could lead to an improvement in the recognition performance. Thus, we adopt attention mechanism to the backbone network in the interaction recognition network to capture more descriminative and powerful feature for recognition. 

\begin{figure}[htbp]
  \centerline{\includegraphics[width = 88 mm]{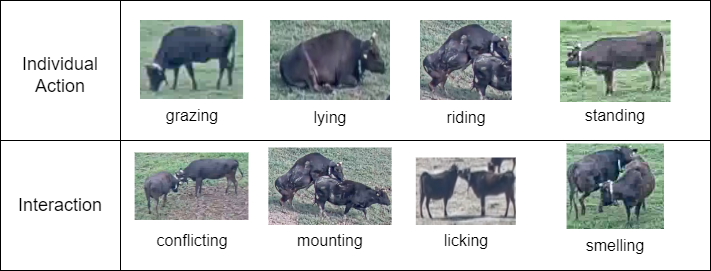}}
    \caption{Classes of cattle individual action and interaction.}
    \label{fig:classes}
\end{figure}
More essentially, previous work has basically been done by recognizing individual action, however, the sociality of cattle, as shown in Fig.~\ref{fig:classes} (interaction), should be concerned to monitor cattle conditions more comprehensively. For example, in addition to estrus there are conflict, communication and others. One of our key observation is that interaction of cattle is related to the individual action. For example, their individual action (`riding' and `standing') and interaction (`mounting') frequently appear together related whereas their individual behavior (`grazing' and `lying') and interaction (`fighting')  almost never happens at the same time \textit{(i.e., the interaction recognition of cattle can be built on the basis of individual action recognition but not two unrelated tasks)}. Our core idea is that based on the extracted feature of individual action, additional features are added and then fused in a mode of late fusion, specifically the spatial relationship between the interaction pair and the visual representation of the interaction region. In addition, we also use the recognized individual action labels as language prior to assist the inference of interaction.

State-of-the-art computer vision systems are usually based on supervised learning models so that these models are critically depend on large datasets requiring enormous human annotation effort. The lack of such large labeled datasets which is problematic for training deep Convolutional Neural Networks (CNNs) due to the overfitting issue, happens to be one of the biggest problems facing the application of computer vision to monitor cattle. This motivates us to look beyond the supervised methods in order to ensure that the models remain valid without the support of large amounts of labeled data. A well-established paradigm in computer vision field has been to pre-train models using large-scale data(e.g., ImageNet\cite{b9}) and then fine-tuned on target tasks with less labeled training data. However, the class of action we want to recognize is very different from the general classification task, so this paradigm is not suitable for our task.

The field of natural language processing (NLP) has made great progress in recent years through learning directly from raw text\cite{b10,b11}. These results suggest that the representation learning ability of modern pre-training methods within extra large scale unlabeled data surpasses that of high-quality labeled NLP datasets. However, in computer vision field is still standard practice to pre-train models on ImageNet or other large scale labeled datasets. Recently, unsupervised visual pre-training has attracted much research attention, which aims to learn a proper visual representation from a large set of unlabeled images. Contrastive learning, which is one type of self-supervised learning, is becoming increasingly attractive due to its great potential to leverage large amount of unlabeled data. The essence of contrastive learning lies in learning 
an embedding space in which similar sample pairs stay close to each other while dissimilar ones are far apart.
Our model is pre-trained in a similar manner to the existing framework SimCLR\cite{b12} on a large unlabeled dataset collected from cattle and then fine-tuned on a relatively small labeled dataset. As a result, both convergence rate and final accuracy of the model are significantly better than random initialization.

Thus, in this paper, we propose a novel framework that can learn to recognize interaction of cattle. The main contributions of this paper is threefold:
\begin{enumerate}
    \item We propose an interaction recognition module based on a still image. Using still images for recognizing saves a significant amount of computation
    . In the proposed module, we construct a triple-stream network to capture different levels of information to represent the feature of interaction. Further, we also adopt attention mechanism to improve the performance of the network.
\item The backbone network in our interaction recognition module is pre-trained on a large dataset of unlabeled cattle data, which is shown be able to learn robust representation of cattle action. 
\item To the best of our knowledge, we are the first to introduce the use of deep learning into the area of animal interaction recognition. And qualitative and quantitative evaluation evidences the performance of our framework as an effective method to recognize cattle interaction.
\end{enumerate}
\section{Related work}
\textbf{Object Detection} is an essential building block for both action and interaction recognition. Most of the recent successful approaches for object detection are based on CNNs. This is due to the automatic feature extraction and the powerful image representation learning ability of the CNN network. Among these approaches there are one-stage and two-stage methods, which are represented by YOLO series\cite{b13,b14,b15} and RCNN series\cite{b16,b17,b18}. Our work uses the YOLOv5 to localize the cattle because of the faster inference speed of one-stage models in order to achieve real-time detection.

\textbf{Action Recognition} has grown more sophisticated with every passing day due to the application of new methods such as two-stream-based\cite{b19}, skeleton-based\cite{b20} and more. However, in terms of two-stream-based methods, which extracting spatial-temporal features from two stream network (RGB flow and Optical flow)\cite{b21,b22} and capturing long-range dependencies through Non-local Neural Networks\cite{b23}, complex architecture results in a huge amount of calculation. As a result, it is difficult to adapt this method to the cattle action recognition because of the large amount of cattle in single frame and the difference between the movement patterns of cattle and human beings
. Indeed, behavior of cattle appears to be very slow compared to that of humans.

Recognition using skeleton data  is generally considered as a time series problem, in which the characteristics of body postures and their dynamics over time are extracted to represent a human action. Although the accuracy is high, it requires special equipment to collect and to create a large number of cattle skeleton dataset which is difficult to be practically implemented.

For the above reasons, we implement the recognition based on still image.

\textbf{Attention} plays an important role in human perception as well as in computer vision models. It makes it possible that a network can weight features by level of importance to a task, and use this weighting to help achieve the task. Extensive efforts have been incorporated in action recognition\cite{b24}. Recent findings in Mobile Network design show that channel attention (e.g., SE attention)\cite{b43} has a significant effect on improving model performance. 
Our work select the Coordinate Attention (CA)\cite{b25}, which embeds positional information into channel attention and performs better than other attention methods with the lightweight property, to improve the performance of our network.   

\textbf{Interaction Recognition} provides a deeper understanding in a scene for highlighting information about the association between objects. Considering the Human-object interactions (HOI) tasks, such tasks are tackled by detecting people doing actions and the object they are interacting with.  Learning the relationship between human and object in a scene, which is called visual relationship learning, with various semantic roles leads to finger-grained understanding of the current activity. Several papers leverage some forms of language prior\cite{b26,b27} to help overcome this task.

In addition to using object instance appearances, Chao et al.\cite{b28} also encode the relative spatial relationship between people and objects with a CNN. Our work draws on these ideas. 

\textbf{Contrastive Learning}, which is one kind of self-supervised learning, has recently narrowed the gap between supervised learning and unsupervised learning. State-of-the-art contrastive learning methods are trained by reducing the distance between representations of different augmented views of the same image (‘positive pairs’), and increasing the distance between representations of augmented views from different images (‘negative pairs’)\cite{b29,b30,b31,b32}. Our approach is inspired by these and our goal is to learn the general representation of cattle's action using an unlabeled cattle dataset.
\section{Method}
\begin{figure}[htbp]
    \centerline{\includegraphics[width=88mm]{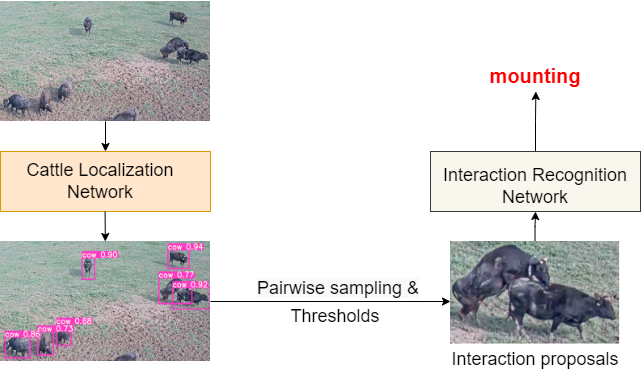} }
    \caption{The pipeline of our proposed method}
    \label{fig:pipeline}
\end{figure}

In this section, we present our method for cattle interaction recognition (see Fig.~\ref{fig:pipeline}). We start with an overview of our approach and then introduce the cattle localization network. Next, we outline the details of the triple-stream interaction recognition network, where the triple-stream refers to the semantic stream, the visual stream and the geometric stream. After that, based on the output of the triple-stream network, we 
explain calculation of interaction prediction score. Finally, we discuss about the supervised training and unsupervised pre-training for our method.
\subsection{Overview}
Our approach to cattle interaction recognition consists of two main steps: 1) cattle detection and 2) interaction recognition. First, given an input image we use YOLOv5 from ultralytics\cite{b33} to detect all the cattle. After getting the bounding box of each cattle, we obtain all of the pair samples with crossover region and then set two thresholds to select high-quality interaction proposals from these samples. The details can be seen in next subsection. When the presence of interaction pair in a current frame is confirmed, we totally obtain four cropped image from the frame as shown in Fig.~\ref{fig:interaction recognition}, two slices of the bounding box area of each cattle, one slice of the interaction area of the two cattle and a two-channel binary image which represent the spatial relationship. Second, we evaluate all the interaction pairs through the proposed interaction recognition network to predict the interaction score as shown in Fig.~\ref{fig:pipeline}.
\subsection{Cattle localization network }
The cattle localization network is mainly build on a YOLOv5 detector and firstly get the bounding box of every cattle. Then we obtain all of the pairwise samples based on whether there is crossover region between two bounding boxes and further set a threshold to filter the mutual obscuration as interaction proposals. Here, we use the  Intersection over Union (IoU)  threshold, which measures the overlap between two bounding boxes.

Let us suppose that two bounding boxes $B_1$ and $B_2$ have coordinates $(x_{1,1},y_{1,1},x_{1,2},y_{1,2})$ and $(x_{2,1},y_{2,1},x_{2,2},y_{2,2})$. Here, $(x_{i,1},y_{i,1})$ denotes the left upper and $(x_{i,2},y_{i,2})$denotes the right lower coordinates of the bounding box. The value of IoU threshold is set as
\begin{equation}
     0.2 < IoU(B_1,B_2) < 0.7 .
     \label{eq:threshold}
\end{equation}
The coordinates of the bounding box of interaction region of $B_1$ and $B_2$, denoted as $B_i(B_1, B_2)$, is defined as
\begin{equation}
\begin{split}
 B_i(x_1,y_1,x_2,y_2) = & (max(x_{1,1},x_{2,1}),max(y_{1,1},y_{2,1}),\\ &min(x_{1,2},x_{2,2}),min(y_{1,2},y_{2,2})). 
\end{split}
\label{eq:intregion}
\end{equation}
If an extracted pair satisfies \eqref{eq:threshold}, then it is used for proposing a final interaction. In our method, the inputs to our interaction recognition network are defined by four slices obtained from interaction proposals, as shown in Fig.~\ref{fig:interaction recognition}, cattle slice $C_1$ and $C_2$, interaction region slice $I$ defined by \eqref{eq:intregion}, and a two channel binary image obtained from two bounding box used in geometric stream.
\subsection{Interaction recognition network}
\begin{figure*}[htbp]
\centering
\includegraphics[width=150mm]{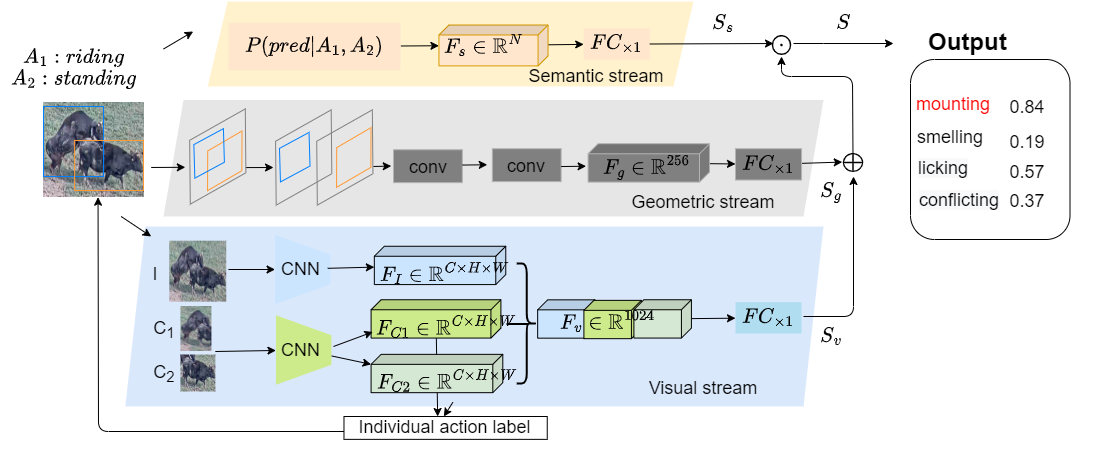}
\caption{Illustration of our triple-stream interaction recognition network.}
\label{fig:interaction recognition}
\end{figure*}

The architecture of the interaction recognition network, as shown in Fig.\ref{fig:interaction recognition}, mainly consists of three streams: 1) the semantic stream 2) the visual stream and 3) the geometric stream. We propose the triple-stream network considering that fusion of different types of features is important for interaction recognition\cite{b32,b34}. For each interaction pair, three types features, \textit{i.e., semantic feature $F_s$, geometric feature $F_g$, and visual feature $F_v$} are extracted from each stream. Then, they are fused to inference the interaction class.

\textbf{Semantic feature $F_s$.} It captures our prior knowledge of cattle behavior. Its motivation is that, in general, the interaction between two cattle is related to their individual action. For example, the interaction between a standing-riding pair is most likely to be mounting, and unlikely to be smelling or conflicting. We build $F_s\in \mathbb{R}^N$ as the frequency of label inter co-occurrence between the interaction and individual action classes\cite{b35}, where $N$ denotes the number of predefined interaction classes in our dataset. The individual action classes we use here are derived from the recognition results of the visual stream. 

More precisely, the semantic feature $F_s$ is obtained by counting the occurrences of predefined interaction classes $pred$ given individual action classes $A_1~and~A_2$. This conditional probability $P$ is calculated as follows:
\begin{equation}
    F_s \in \mathbb{R}^N  = P(pred|A_1,A_2)
\end{equation}
where $pred$ refers to each interaction class. 

\textbf{Geometric feature $F_g$.} To characterize the spatial relationship between the interaction pair, we adopt the two-channel binary image representation in\cite{b28}. Specifically, we first take the union of the two boxes and construct a binary image with two channels and pixels in each channel has value $1$ within the bounding box area and value $0$ elsewhere. We then use two convolution layers and one pooling layer to extract spatial feature $F_g \in \mathbb{R}^{256}$ from the two-channel binary image.

\textbf{Visual feature $F_v$.} Compared with the  $F_g$ and  $F_s$, it is supposed to play the most important role in the recognition of interaction. As shown in Fig.~\ref{fig:interaction recognition}, cattle slice $C_1$, $C_2$ and interaction slice $I$ are input into the visual stream and we use two CNNs to extract both individual action feature of $C_1$, $C_2$ and interaction feature of $I$. A concatenation of the three features will be used as the final visual feature $F_v$. 

Here, both of the two CNNs is based on the architecture of EfficientNet\cite{b36}, which is a kind of mobile network with extremely high efficiency and speed. The network uses all dimensions of the recombination coefficient unified scaling model to greatly improve the accuracy and efficiency of the model. To further improve the feature extraction capability of the network, we adopt coordinate attention (CA)\cite{b25} to the network. The ability to encode horizontal and vertical location information into channel attention allows mobile networks to focus on a large range of location information without imposing too much computational effort. The CA is added to the residual blocks of the EfficientNet. See \cite{b25} for the details on CA.

Let us denote the height and the width of images, and the number of channels of images as $H, W, C$ respectively. The obtained three features are denoted as $F_{C1} \in  \mathbb{R}^{C \times H \times W} $, $F_{C2} \in \mathbb{R}^{C \times H \times W}$ and $F_I \in \mathbb{R}^{C \times H \times W}$. Based on these, the final visual features are represented as follows: 
\begin{equation}
  F_v = [F_{C1} \cdot F_{C2} \cdot F_I] \in \mathbb{R}^{3C \times H \times W},
  \label{equ:visual feature}
\end{equation}
where $[\cdot]$ refers to the concatenation operation. At the same time, the individual feature $F_{C_1}$ and $F_{C_2}$ shown in visual stream in Fig.~\ref{fig:interaction recognition} will be separately used to predict the individual action classes $A_1$ and $A_2$, which are used in the semantic stream, through an FC layer and a softmax layer.
\subsection{Calculation of interaction prediction score}
Through localization network, only a few high-quality cattle pairs are supposed to be preserved and fed into the triple-stream interaction recognition network. For each interaction pair candidate, the semantic $F_s$, geometric $F_g$ and visual $F_v$ will be summed in a late fusion way, which means that the final prediction scores are independently predicted from the triple stream and then summed later. These three scores are defined as follows:
\begin{equation}
\begin{split}
    score~for~semantic~stream : S_s &= \sigma(FC_{\times1}(F_s)) \in [0,1]^N,\\
    score~for~geometric~stream : S_g &= \sigma(FC_{\times1}(F_g)) \in [0,1]^N,\\
    score~for~visual~stream : S_v &= \sigma(FC_{\times1}(F_v)) \in [0,1]^N,
\end{split}
\end{equation}
where $S_s, S_g, S_v$ are the score vector from triple streams, respectively, and $\sigma$ refers to ReLu function. 

During inference, the final prediction score is calculated as follows:
\begin{equation}
S = (S_v + S_g) \odot S_s,
\label{equ:score fusion}
\end{equation}
where $\odot$ denotes the Hadamard product. 
In this subsection, we introduce the training process of our proposed method. It is divided into two parts. One is supervised training of the entire framework and the other is self-supervised pretraining for the two CNNs in the Interaction Recognition Network. Due to the small amount of labeled data, random initialization of the weights of the two CNNs  may lead to overfitting, so we adopt transfer learning to use the pretrained model instead of random initialization and then fine-tune it during supervised training. 

\textbf{Supervised training}. The localization network and the interaction recognition network of our architecture can be trained in an end-to-end manner. The entire loss $\mathcal{L}_{entire}$ is computed as
\begin{equation}
    \mathcal{L}_{entire}=\mathcal{L}_{loc} + \mathcal{L}_{ind} + \mathcal{L}_{int}.
    \label{eq:loss-entire}
\end{equation}
Here, $\mathcal{L}_{loc}$ is the localization loss which is the same as that used in YOLOv5\cite{b33}. $\mathcal{L}_{ind}$ and $\mathcal{L}_{int}$ are both cross-entropy losses in interaction recognition network. $\mathcal{L}_{ind}$ is used to evaluate loss between predicted outcomes and ground truth for individual action classes. Similarly, $\mathcal{L}_{int}$ is used to evaluate loss between final prediction score $S$ and the ground truth for interaction classes. Note that each individual action and interaction class is supposed to be independent and not mutually exclusive. 

\begin{figure}
\centerline{\includegraphics[width=88mm]{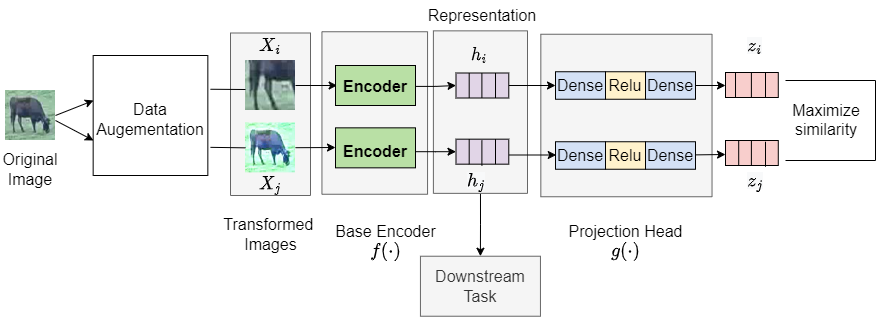} }
\caption{Pre-training framework of our model.}
\label{fig:pretrain}
\end{figure}

\textbf{Self-supervised Pretraining.} Following the idea in Chen et at.\cite{b12}, the method we pretrain the two CNNs is shown in Fig.~\ref{fig:pretrain}. First, pre-trained samples are obtained from the YOLOv5 detection model. We save image slices of all detected cattle to be used as independent samples without annotation. Then, two data argumentation operations, random crop and color distort\cite{b37}, are adopted. Suppose that the batch size is $M$, given two augmented images $X_i$ and $X_j$ from one cattle slice as positive pair, the other $2M-2$ augmented samples are treat as negative samples. The representation $z \in \mathbb{R}^{128}$ of all input image $X$, which will be used to calculate the contrastive loss, can be represented as
\begin{equation}
    z=g(f(X)),
    \label{eq:representation}
\end{equation}
where $f(\cdot)$ denotes the encoder, specifically EfficientNet, and $g(\cdot)$ denotes a linear projection which maps $f(X)$ to the space where contrastive loss is computed, as shown in Fig.~\ref{fig:pretrain}. The learning goal is to make representation of positive samples $z_i$ similar to $z_j$, where $z_i$ and $z_j$ are obtained from original input image $X$ by adopting the calculation defined in \eqref{eq:representation}, and at the same time different from other representation of the rest image of the mini-batch. 
To evaluate the similarity between features, $sim(z_i,z_j)=z_i^\top z_j/ \|z_i\|{_2}\|z_j\|{_2}$ denotes the dot product between ${L}^2$ normalized between $i$ and $j$ (\textit{i.e.cosine similarity}). To distinguish from the loss function of supervised learning, we put the tilde for symbols denoting the pretraining loss in the following. The loss function for a positive pair  $(X_i,X_j)$ can be written as
\begin{equation}
    \tilde{\mathcal{L}}(X_i,X_j) = -\log\frac{\exp(sim(z_i,z_j)/\mathcal{T})}
    {\sum_{k=1}^{2M}\exp(sim(z_i,z_k))}\quad (k\neq i), \label{eq:nt-xent}
\end{equation}
where $\mathcal{T}$ denotes a temperature parameter which scales the input and expands the range of cosine similarity. Note that $k \neq  i$ means $k$ and $i$ should be different samples in the current batch. Equation \eqref{eq:nt-xent} is called as NT-Xent, meaning the normalized temperature-scaled cross entropy.
Finally, we calculate the loss $\tilde{\mathcal{L}}_{all}$ on all pairs in a batch of size $M$ and take the average value, where $\tilde{\mathcal{L}}_{all}$ is defined as
\begin{equation}  
    \tilde{\mathcal{L}}_{all}=\frac{1}{2M}\sum^M_{k=1}\tilde{\mathcal{L}}(X_{2k-1},X_{2k})+ \tilde{\mathcal{L}}(X_{2k},X_{2k-1}),
\end{equation}
where $(X_{2k},X_{2k-1})$ denotes every positive pair of the mini-batch and note that every sample will be calculated twice. The details of pre-training will be given in next section. Our pre-trained model, the encoder shown in Fig.~\ref{fig:pretrain}, will be used in the visual stream by adopting transfer learning to initialize the parameters of the two CNNs. Then the two CNNs will be fine-tuned during the supervised training.
\section{Experiment} 
\subsection{Dataset}               
\begin{figure}[!htbp]
    \centering
    \includegraphics[width=88mm]{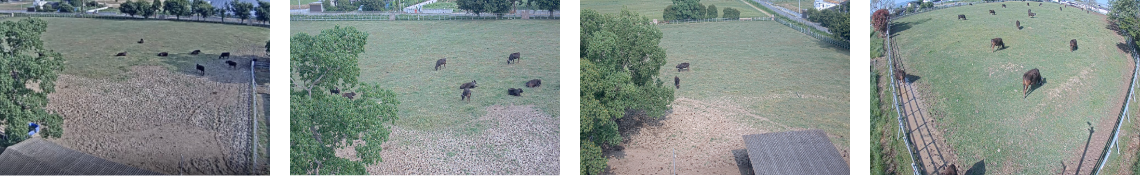}
    \caption{Sensor setting for the experiment.}
    \label{fig:snapshot}
\end{figure}

We present our results on a cattle dataset conducted at a anonymous pasture. About 40 black breeding cattle are grazed and raised in the pasture. An automatic capture system for videos of cattle using several RGB-cameras is set there. These cameras are equipped with a wide-angle lens to capture the whole pasture and they are installed at different angles to capture images from various viewpoints (see Fig.~\ref{fig:snapshot}). This will avoid partial occlusion and increase the amount of data.

For the YOLOv5 detector, the dataset for detection consists of 800 images, of which 600 are used as the training set and the rest are set as a test set. Then, the cattle-individual-action dataset contains 5000 images obtained from the output of the detector. Among the 5000 images, only 1500 samples have annotation and the rest 3500 samples are used for unsupervised pretraining. 
Finally, we have a cattle-interaction dataset which contains 300 labeled images. This dataset is relatively small, mainly because significant interactions occur infrequently and it is very labor-intensive to extract these data from the massive amount of videos. Classes of both cattle-individual-action dataset and cattle-interaction dataset are shown in Fig.~\ref{fig:classes}.
\subsection{Implementation details}
\textbf{Settings for Detector.} Since YOLOv5 is an anchor-based detector and the prior anchor of YOLOv5 is obtained from COCO\cite{b38}, which is not applicable to our work. Thus, the K-means clustering algorithm is adopted to cluster width and height of the annotated bounding boxes. YOLOv5 is implemented with a feature backbone of Darknet\cite{b14} and neck with PAN\cite{b39} and FCN\cite{b40}. We keep cattle boxes with scores higher than $0.7$.

\textbf{Settings for Unsupervised pretraining.}  Unless specially noted, we adopt the following settings for pretraining. Following the conclusion raised in\cite{b12} that contrastive learning benefits from larger batch size, we resize all the 5000 training samples to $100\times100$ pixels, and we train the model on three NVIDIA P100 GPU in a batch size of $512$. We alternatively random crop and color distortions for data augmentation. We use Efficient-B0\cite{b36} with coordinate attention\cite{b25} as base encoder, and a two-dense-layer projection head shown in Fig.~\ref{fig:pretrain} to map the representation extracted by base encoder to a 128-dimensional latent space. Then we optimize the NT-Xent loss (\ref{eq:nt-xent}) using Adam with learning rate of $0.4$ and weight decay of $10^{-6}$. The temperature parameter $\mathcal{T}$ is set as $0.5$. The pretrained network is used in our interaction recognition network for initializing the parameter of the two CNNs in visual stream.

\textbf{Settings for supervised training.} The learning goal of supervised training is minimize the entire loss $\mathcal{L}_{entire}$ \eqref{eq:loss-entire}. As mentioned above, the pretrained model is transferred to initialize the two CNNs visual stream and we adopt transfer learning in two modes:
\begin{enumerate}
    \item \textbf{Transfer learning via a linear classifier.} In this way, we train  just the fully-connected layer on feature extracted from the frozen pretrained network.
    \item  \textbf{Transfer learning via Fine-tuning.} We fine-tune the entire network using the weights of pretrained network as initialization.
\end{enumerate}

Note that each of the two modes uses the same procedure and the two convolution layers in geometric stream are always randomly initialized. We train for 500 steps at a batch size of 128 using Adam to optimize the softmax cross-entropy objective. The learning rate and weight decay are set to be the same as in unsupervised pretraining. To demonstrate the effectiveness of transfer learning, we also compare the performance of transfer learning and learning with random initialization.
\subsection{Quantitative evaluation}
\textbf{Cattle interaction detection results.} We present the overall quantitative results using the mean Average Precision ($mAP$)\cite{b41} as evaluation metric. For cattle-interaction dataset, since there is no existing method for detecting cattle interaction, we compare our method with general detection framework Faster RCNN, YOLOv5, EfficientDet and report the quantitative evaluation of interactions in Table~\ref{table:map}. In comparison with other detector models, our method improves the $mAP$ by 12.6\% more than the highest of the existing detection framework.
\begin{table}[htbp]
\begin{center}
\caption{Performance comparison with original Faster RCNN, YOLOv5, and EfficientDet on cattle-interaction test set.}
\label{table:map}
\begin{tabular}{ll|l}
\hline\noalign{\smallskip}
\bf Method $\qquad\qquad$& \bf Feature backbone & \bf $mAP$\\
\noalign{\smallskip}
\hline
\noalign{\smallskip}
Faster RCNN  & Resnet101-FPN $\qquad$& 35.5\\
YOLOv5 & Darknet & 33.9\\
EfficientDet-D0 & EfficientNet-B0 + BiFPN  & 42.1\\
Ours & EfficientNet-B0 & 49.3\\ 
Ours & EfficientNet-B0 + CA  & \textbf{54.7}\\
\hline
\end{tabular}
\end{center}
\end{table}

\begin{figure}[htbp]
    \centering
    \includegraphics[width=80mm]{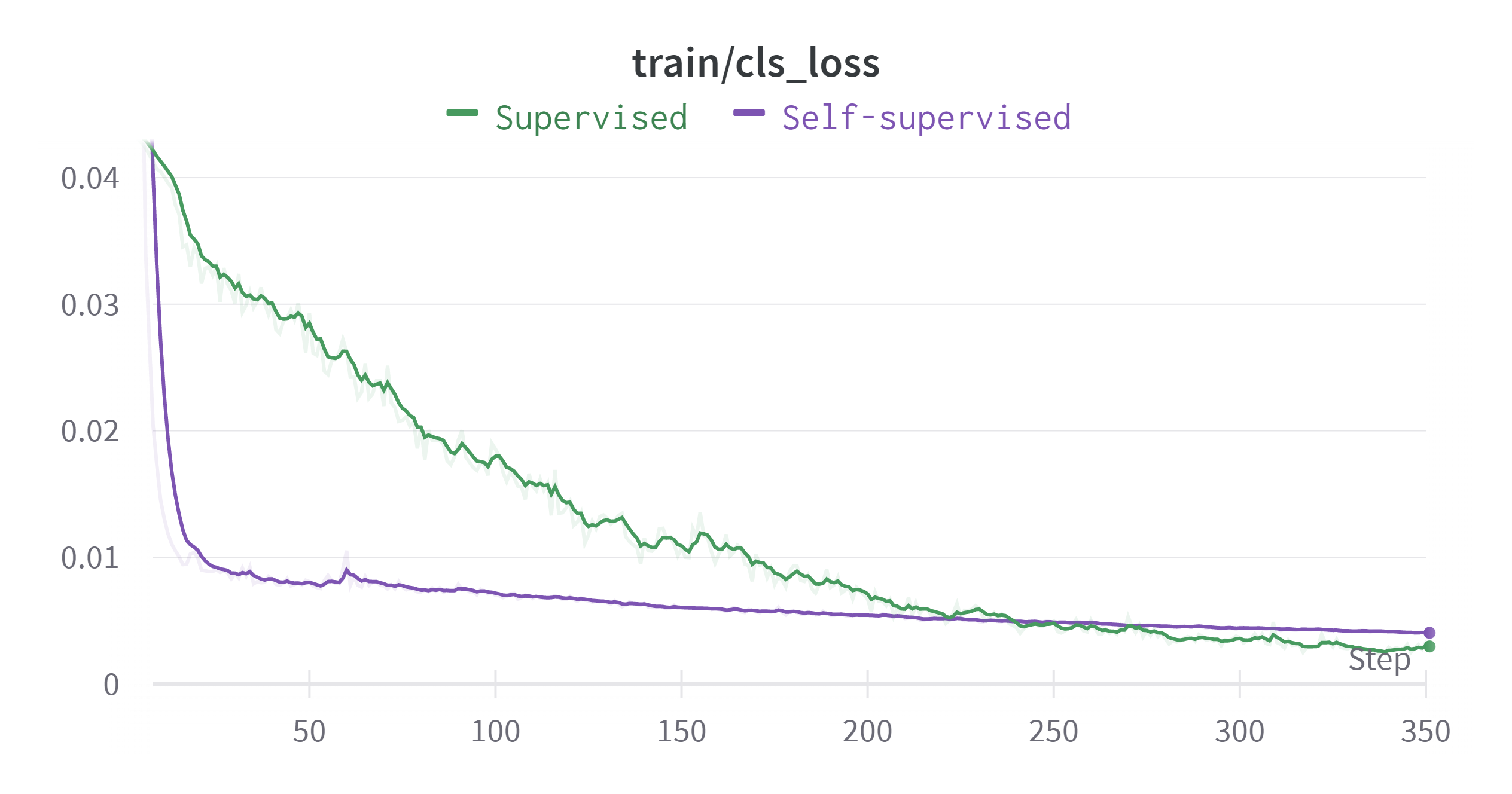}
    \caption{Comparison of convergence rate between self-supervised and supervised methods.}
    \label{fig:convergence}
\end{figure}

\textbf{Evaluation of self-supervised pretraining.} We investigated three methods of transfer learning performance on the cattle-individual-action dataset and compared the accuracy with a supervised baseline model with random initialization using the same backbone network with standard cross-entropy loss. The accuracy is reported on the test set as shown in Table~\ref{table:accuracy}. The results demonstrate that the transfer learning via fine-tuning achieves the best accuracy between the two types of transfer learning and even has better performance compared to the supervised model. In addition, we also report the convergence rate between the supervised and the fine-tuned self-supervised model in Fig.~\ref{fig:convergence}. According to the results, the convergence rate of the self-supervised model is shown to be faster.

\begin{table}[htbp]
\caption{Comparison of performance of our self-supervised approach with supervised baseline.}
\centering
\label{table:accuracy}
\begin{tabular}{l|ll}
\hline\noalign{\smallskip}
\bf{Method} & \bf{Accuracy}\\
\noalign{\smallskip}
\hline
\noalign{\smallskip}
Linear Classifier& 57.39 &   \\
Fine-Tuning & \textbf{71.21} & \\
     Random Initialization & 63.18 & \\
\hline 
\end{tabular}
\end{table}

\textbf{Ablation study.} Since the final prediction of interaction is predicted by three features, visual feature $F_v$, semantic feature $F_s$ and geometric feature $F_g$, we adopt key component analysis to clarify the efficacy of these three features. Table~\ref{tab:ablation} reports the performance with considering one single feature or two. As seen, though the visual feature plays a more important role than the other, the other two features also show their contribution to the final prediction. In addition, we also investigate different ways of score fusion from the three features, the results demonstrate that the one used in (\ref{equ:score fusion}) is the best. 
\begin{table}[htbp]
    \centering
    \caption{Ablation study of interaction representation and score fusion.}
    \begin{tabular}{l|l|l}
    \hline\noalign{\smallskip}
    \textbf{Aspect}    & \textbf{Variant} & \textbf{$mAP$}  \\
    \hline
    \multirow{5}{*}{\makecell[l]{Interaction\\  Representation}}  & only $F_s$  & 34.5 \\
     & only $F_g$ & 9.6 \\
     & only $F_v$ & 43.7 \\
     &$F_v + F_g$ & 46.8 \\
     &$F_v + F_s$ & 51.1 \\
     & $F_v + F_g + F_s$ & \textbf{54.7} \\
     \hline 
     \multirow{3}{*}{\makecell[l]{Score \\ Fusion}} & $S_v +S_g +S_s$ $\qquad$  &50.6  \\
     &$S_v  \odot S_g  \odot S_s$ & 51.2\\
     &$(S_v  + S_g)  \odot S_s$ & \textbf{54.7} \\
     \hline
    \end{tabular} 
    \label{tab:ablation}
\end{table}
\subsection{Qualitative evaluation}
\begin{figure}[htbp]
    \centering
    \includegraphics[width=88mm]{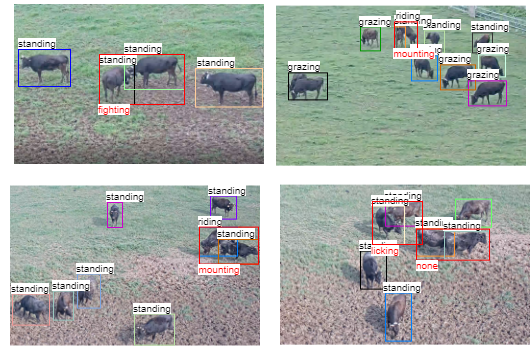}
    \caption{Sample detections on the cattle interaction test set.}
    \label{fig:sample of result}
\end{figure}

\begin{figure}[htbp]
    \centering
    \includegraphics[width =88mm]{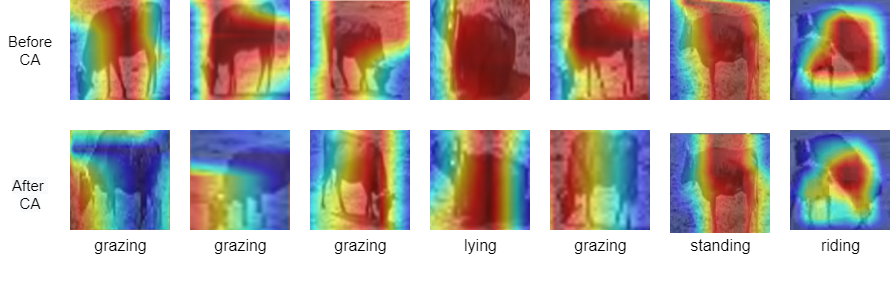}
    \caption{Comparison of feature maps before and after coordinate attention.}
    \label{fig:Attention}
\end{figure}

\textbf{Cattle interaction detection results.} Here, we give out some samples of interaction detection results as shown in Fig.~\ref{fig:sample of result}. We highlight the detected interaction with red bounding box. The results show that our framework is able to accurately locate the interaction pairs in a herd and recognize the class of interaction. In addition, our system's processing speed is roughly 20fps, and because cattle usually move slowly, there is no need to process every frame, so we set recognition every five frames to achieve the purpose of real-time monitoring
\\
\textbf{Attention map visualization.} Fig.~\ref{fig:Attention} visualizes the feature map produced by models with coordinate attention methods in the last building block. Grad-CAM\cite{b42} is used as our visualization tool. Obviously, the coordinate attention can help better in locating the distinctive area of the action which is consistent with our observations noted in introduction. 
\section{Conclusions}
In this paper, we present a novel framework for detect interaction between cattle in a pasture. Our core idea is to fuse the features of individual regions, interaction regions, and spatial relationships, and to further utilize the recognition results of individual actions for a prior knowledge of interaction inference. We quantitatively evaluate the effectiveness of each component of the method and the performance improvement over a common detection framework. This is the first interaction detection framework applied to cattle. We believe this research will help agriculturalists to manage and raise cattle more easily and will continue to study it further. 

\end{document}